\documentclass[10pt,twocolumn,letterpaper]{article}
\PassOptionsToPackage{table}{xcolor}
\usepackage[pagenumbers]{cvpr} 

\usepackage{pifont} 

\definecolor{cvprblue}{rgb}{0.21,0.49,0.74}
\usepackage[pagebackref,breaklinks,colorlinks,allcolors=cvprblue]{hyperref}
\usepackage{multirow}
\usepackage{amssymb}
\usepackage{pifont}
\usepackage{array}
\usepackage{booktabs} 
\usepackage{float}
\usepackage{graphicx}

\def\confName{CVPR}
\def\confYear{2026}

\title{ChronusOmni: Improving Time Awareness of Omni Large Language Models}

\author{Yijing Chen$^{1\dagger}$, Yihan Wu$^{1\dagger}$, Kaisi Guan$^{1}$, Yuchen Ren$^{1}$, Yuyue Wang$^{1}$, Ruihua Song$^{1*}$, Liyun Ru$^{2*}$\\
$^1$Gaoling School of Artificial Intelligence, Renmin University of China \ 
$^2$Baichuan Inc.\\
}

\newcommand{\ModelName}{ChronusOmni}
\begin{document}
\maketitle
\begin{abstract}
Time awareness is a fundamental ability of omni large language models, especially for understanding long videos and answering complex questions. Previous approaches mainly target vision-language scenarios and focus on the explicit temporal grounding questions, such as identifying when a visual event occurs or determining what event happens at a specific time. However, they often make insufficient use of the audio modality, and overlook implicit temporal grounding across modalities—for example, identifying what is visually present when a character speaks, or determining what is said when a visual event occurs—despite such cross-modal temporal relations being prevalent in real-world scenarios.
In this paper, we propose \textbf{ChronusOmni}, an omni large language model designed to enhance temporal awareness for both explicit and implicit audiovisual temporal grounding. First, we interleave text-based timestamp tokens with visual and audio representations at each time unit, enabling unified temporal modeling across modalities. Second, to enforce correct temporal ordering and strengthen fine-grained temporal reasoning, we incorporate reinforcement learning with specially designed reward functions. Moreover, we construct \textbf{ChronusAV}, a temporally-accurate, modality-complete, and cross-modal-aligned dataset to support the training and evaluation on audiovisual temporal grounding task.
Experimental results demonstrate that \ModelName~achieves state-of-the-art performance on ChronusAV with more than 30\% improvement and top results on most metrics upon other temporal grounding benchmarks. This highlights the strong temporal awareness of our model across modalities, while preserving general video and audio understanding capabilities. Code and dataset are available at \url{https://github.com/YJCX330/Chronus/}.

\renewcommand{\thefootnote}{\fnsymbol{footnote}}
\makeatletter
\renewcommand{\@makefntext}[1]{%
  \parindent 1em\noindent
  \hbox to 1.8em{#1\hss}} %
\makeatother
\footnotetext[1]{$^\dagger$Equal contribution. Contact chenyijing@ruc.edu.cn}
\footnotetext[2]{$^*$Corresponding authors.}

\end{abstract}    
\section{Introduction}
\label{sec:intro}
\begin{figure}[t]
  \centering
  \setlength{\abovecaptionskip}{1.0mm}
   \includegraphics[width=1.0\linewidth]{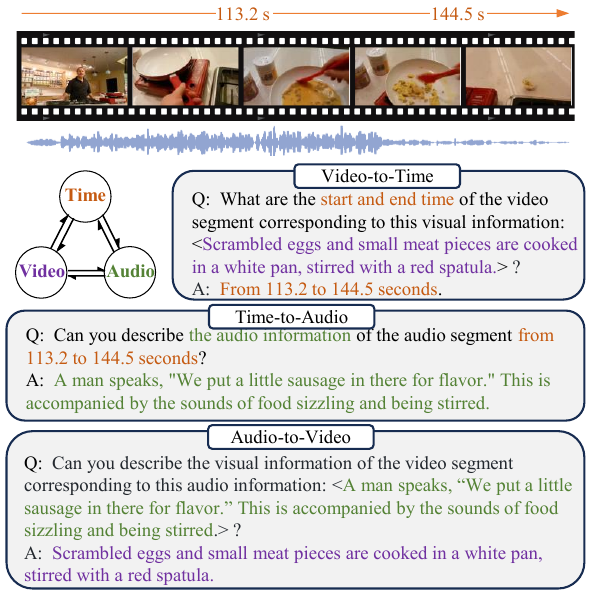}
   \caption{Illustration of the audiovisual temporal grounding task. Three primary elements Video, Audio and Time are connected through six directional basic temporal grounding subtasks: Video-to-Time, Time-to-Video, Audio-to-Time, Time-to-Audio, Video-to-Audio, and Audio-to-Video.}
   \label{fig:teaser}
     \vspace{-6mm}
\end{figure}

Video temporal grounding~\cite{vtgsurvey,guo2025trace,guo2025vtgllm} aims to align video semantics with temporal segments, serving as a critical capability for video understanding. 
Early studies in this field~\cite{HawkEye,guo2025trace,guo2025vtgllm,huang2024lita,timechat,chen2024timemarker,huang2024vtimellm,timer1} primarily rely solely on visual information to align video content with textual descriptions. However, as in-the-wild videos increasingly contain rich audio cues (dialogues, environmental sounds, and background music), visual information alone is insufficient for precise grounding. This motivates a shift toward audiovisual temporal grounding, which incorporates both visual and auditory modalities to capture complementary cues and improve time awareness in complex real-world scenarios.

Temporal grounding serves as a prerequisite for sophisticated multimodal reasoning.
Consider a typical audio-visual scenario (e.g., in a movie dialogue scene, the camera first stays on Character A’s expression; off-screen, Character B delivers the line “Let’s go now.” Then the shot cuts to B, and B’s lip movements align with the dialogue.), 
where the semantic meaning of an event can only be accurately understood when both visual and auditory cues are jointly analyzed and temporally aligned. 
In such cases, it requires model to understand not only the \textbf{localization information in video streams} and the \textbf{localization information in audio streams} (explicit temporal grounding), but also \textbf{the synchronization of audio-visual interactions} (implicit temporal grounding). 
However, most existing works focus solely on explicit temporal grounding
~\cite{guo2025trace,Zeng2025videochatt,huang2024vtimellm,huang2024lita,vidi}, relying on either visual frames or audio features without capturing temporal dependencies cross audio and video.
As a result, they struggle in complex real-world scenarios where auditory and visual cues complement each other. This limitation reveals a fundamental challenge for multimodal temporal grounding: models must be capable of jointly representing, aligning, and reasoning over temporal dynamics across both auditory and visual streams. 
To address the above challenge, in this work, we improve the audiovisual temporal grounding task in a systematic way. First, we formally define the audiovisual temporal grounding task and systematically analyze its core challenges. As illustrated in Figure~\ref{fig:teaser}, a model with audiovisual temporal grounding capability should support six directions of prediction across time, video, and audio:\textbf{ video-to-time}, \textbf{time-to-video}, \textbf{audio-to-time}, \textbf{time-to-audio}, \textbf{audio-to-video}, and \textbf{video-to-audio}. Among them, the first four directions are explicit temporal grounding, while the last two directions constitute implicit temporal grounding.
This requires three essential abilities: (1) video content localization, (2) audio content localization, and (3) synchronization modeling of audio-visual interactions. Second, we design audiovisual temporal-synchronized representation, improving multimodal joint temporal reasoning ability with temporal-aware training strategy. Furthermore, we construct a comprehensive audiovisual dateset to support the training and evaluation for audiovisual temporal understanding.

The main contributions can be summarized as follows.
\begin{itemize}[leftmargin=1em]
    \item We propose \textbf{\ModelName}, an omni large language model designed for the audiovisual temporal grounding task. More specifically, we design an audiovisual temporal interleaved representation that aligns visual, audio, and temporal information effectively. We further design the temporal-aware supervised finetuning and reinforcement learning strategy upon the representation, showing remarkable performance on both audiovisual joint temporal understanding and unimodel understanding.
    \item We introduce \textbf{ChronusAV}, a temporally-accurate, modality-complete, and cross-modal-aligned dataset to support the training and evaluation on the audiovisual temporal grounding task, covering six cross-modal prediction directions among time, video, and audio.
    \item Experimental results show that \ModelName~achieves the best performance in all metrics over ChronusAV (with more than 30\% improvement) and most metrics over other temporal grounding benchmarks, while preserving its general video and audio understanding capabilities.
    These results demonstrate the effectiveness of our proposed temporal modeling method and training strategy.
\end{itemize}

\section{Related Work}
\label{sec:relatedworks}

\begin{figure*}[t]
\centering
\includegraphics[width=1.0\linewidth]{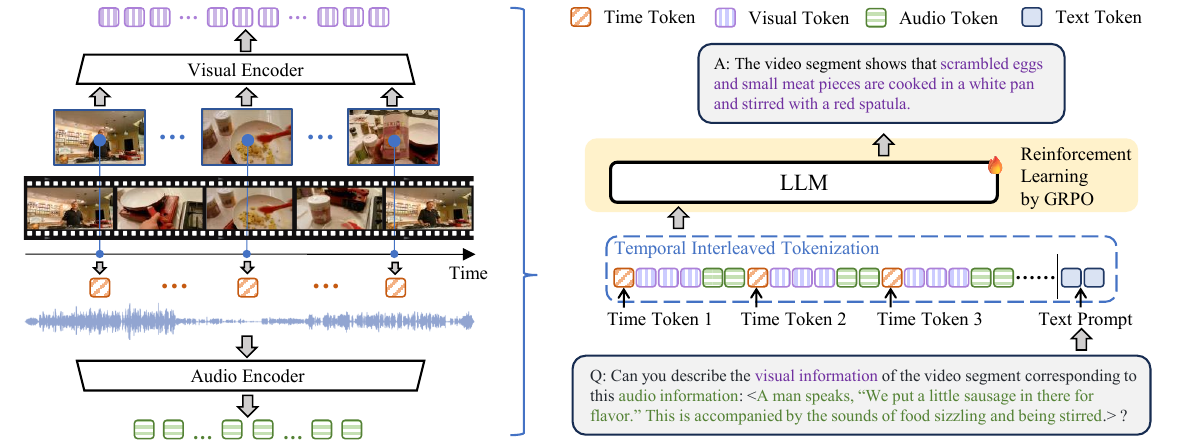} 
\vspace{-1.5 em}
\caption{The architecture of ChronusOmni. Time, video, audio are tokenized and interleaved at each time step. The token sequence is along with text prompt is input into an LLM, which is supervised finetuned and further enhanced by reinforcement learning.}
\label{fig:architecture}
\vspace{-1.5 em}
\end{figure*}

\paragraph{Temporal Grounding MLLMs.} Traditional video temporal grounding~\cite{vtgsurvey} aims to precisely align video semantics with corresponding time, including various tasks such as moment retrieval~\cite{lisa2018mr,gao2017mr,zala2023mr,boris2024mr}, dense video caption~\cite{yang2023vid2seq,kim2024dvc}, and video highlight detection~\cite{vhd}. With the development of video large language models (LLMs), some recent works leverage the reasoning capabilities of LLMs to enhance temporal grounding. Some models directly infer frame indices relying on their ability to reason over frame order~\cite{huang2024vtimellm,videochatflash}, while others introduce absolute timestamp encoding to strengthen temporal localization~\cite{chen2024timemarker,qian2024momentor,guo2025trace,guo2025vtgllm}.
Beyond the video modality, several audiovisual LLMs~\cite{videollama, qwen25omni,qwen3omni,avicuna,longvale,trisense,hunyuan} extend temporal perception capabilities to audiovisual scenarios.
Many use learnable temporal positional embeddings~\cite{videollama, qwen25omni,qwen3omni} or specially designed time encoders~\cite{trisense}, which require substantial amounts of temporal training data to develop fine-grained temporal sensitivity from scratch. 
ARC-Hunyuan-Video~\cite{hunyuan} instead embeds a watermark representing absolute time on each frame. It introduces additional complexity because the model must perform OCR-like operations to recover temporal information.
To address these limitations, we propose an audiovisual temporal-synchronized representation. We encode time directly as a text modality and interleave audio, visual, and time tokens along the timeline, enabling explicit, fine-grained alignment across modalities. With our temporal-aware training strategy, the model acquires strong audiovisual temporal grounding ability using only a relatively small amount of training data.
\vspace{-5mm}
\paragraph{Temporal Grounding Datasets.}
Most existing temporal grounding benchmarks~\cite{gunnar2016charades,gunnar2016charades,activitynet,zhou2018youcook2,lisa2017didemo,lei2020tvr} rely on visual information solely, neglecting the audio modality entirely. AVEL~\cite{tian2018AVEL} and UnAV-100~\cite{geng2023UnAV100} incorporate both audio and video streams with coarse event-level labels without detailed captions, limiting the model's ability to understand fine-grained audiovisual temporal details. LongVALE~\cite{longvale} and TriSense-2M~\cite{trisense} offer more detailed temporal event descriptions that combine information from both modalities without separate audio and video captions. This entanglement makes it difficult to analyze or assess a model’s temporal understanding capability within and across each modality independently. VUE-TR~\cite{vidi} provides separate temporal annotations for video and audio streams, but its audio annotations focus only on human speech, restricting its applicability to more diverse audio-visual scenarios. Furthermore, none of the existing temporal grounding benchmarks evaluate implicit temporal perception across audio and video modalities, such as grounding what is heard when something is seen, or what is seen when something is heard.
To address these limitations, we introduce ChronusAV, a temporally-accurate, modality-complete, and cross-modal-aligned dataset that provides time aligned triplets of (timestamp, video segment caption, audio segment caption) for each video. This design enables flexible construction of all six multimodal temporal grounding subtasks (as shown in Figure~\ref{fig:teaser}) and establishes a comprehensive foundation for advancing multimodal temporal grounding research.

\section{Task Definition: Audiovisual Temporal Grounding}

In this section, we formally define the audiovisual temporal grounding task and its six subtasks accordingly.
Suppose we have a video $v$ and its corresponding audio $a$, both with a duration $t$. We denote a seqence of absolute timestamps as $t_0,t_1,t_2,...,t_n$, which divide the video into $n$ temporal segments: $(t_0,t_1),(t_1,t_2),...,(t_{n-1},t_n)$. For each interval $(t_{i},t_{i+1})$, we extract a video clip $v_{i}$ and an audio clip $a_i$. The clip $v_i$ (and similarly $a_i$) starts at $t_i$ and ends at $t_{i+1}$. Thus, we obtain $n$ aligned tuples:
\vspace{-2mm}
\begin{equation}
    D_i = \{ v_i, a_i, (t_i, t_{i+1}) \},
    \vspace{-2mm}
\end{equation}
where each tuple contains a video segment, an audio segment, and its corresponding timestamp.

Given a tuple $D_{i}$, the audiovisual temporal grounding task consists of querying one element to infer another. This give rise to six subtasks, as illustrated in Figure~\ref{fig:teaser}.

\begin{itemize}[leftmargin=1em]
    \item \textbf{Video-to-Time} (V2T). Given the information of the video segment $v_i$, the goal is to ground its corresponding absolute time segment $(t_i,t_{i+1})$. 
    \item \textbf{Time-to-Video} (T2V). Given the absolute time segment $(t_i,t_{i+1})$, the objective is to obtain the information of the associated video segment $v_i$.
    \item \textbf{Audio-to-Time} (A2T). Given the information of the audio segment $a_i$, the goal is to ground its corresponding absolute time segment $(t_i,t_{i+1})$. 
    \item \textbf{Time-to-Audio} (T2A). Given the absolute time segment $(t_i,t_{i+1})$, the objective is to obtain the information of the associated audio segment $a_i$.
    \item \textbf{Video-to-Audio} (V2A). Given the visual content of $v_i$, V2A aims to identify the content of the time-synchronized audio segment $a_i$.
    \item \textbf{Audio-to-Video} (A2V). Given the audio information of $a_i$, the goal is to identify the time-synchronized video segment $v_i$.
    
\end{itemize}

\section{Proposed Method: \ModelName}

To equip the model with fine-grained audiovisual time awareness, we propose \ModelName, a unified framework that achieves precise alignment across video, audio, and time. Unlike prior MLLMs that process audio and video largely in isolation, \ModelName~organize tokens in an interleaved manner to explicitly model cross-modal temporal dependencies. In addition, we design a two-stage, coarse-to-fine temporal optimization strategy to strengthen the model’s temporal reasoning ability, resulting in robust performance on multimodal temporal grounding tasks.

\subsection{Audiovisual Temporal Representation}
\label{temporal_modeling}
To improve cross-modal synchronization and achieve fine-grained alignment of audiovisual information and absolute time, we propose a temporal interleaved tokenization method, explicitly aligning the three elements--video, audio, and time along the timeline, as depicted in Figure~\ref{fig:architecture}.

For visual input, we uniformly sample a fixed number of video frames. A visual encoder and audio encoders are used to transform the video frames and audio into modality-specific tokens.
Because the time intervals between frames are not fixed, explicit timestamps are essential for conveying absolute temporal information to the model. For each sampled frame, we extract its corresponding time point and convert it into text following the fixed format \texttt{second\{t\}} in~\cite{chen2024timemarker}.
For example, for a 126-second video sampled at 2-second intervals, we collect 64 timestamps: ``second\{0.0\}", ``second\{2.0\}", ``second\{4.0\}", ..., ``second\{126.0\}". These timestamps are encoded directly into text tokens, making them fully interpretable by the LLM. In contrast to methods that rely on learnable temporal embeddings, our explicit textual timestamps do not require aligning a learned embedding space with the language space, resulting in better adaptability, easy extensibility, and more stable temporal grounding performance.

Before feeding the tokens into the LLM, we interleave the token sequences from these three elements to achieve fine-grained temporal alignment. The interleaved tokens can be represented as:
\vspace{-2mm}
\begin{equation}
    I  = [T_1, V_1, A_1, T_2, V_2, A_2, ..., T_i, V_i,  A_i, ...],
\vspace{-2mm}
\end{equation}
where \( T_i \) is the absolute timestamp tokens corresponding to the specific time point \( t_i \), \( V_i \) is the tokens encoded from the frame at \( t_i \), and \( A_i \) is the tokens encoded from the audio between \( t_i \) and \( t_{i+1} \). By using interleaved tokens, the model can establish a fine-grained connection among absolute time, visual information, and audio information.

\subsection{Audiovisual Temporal Optimization}
\label{optimization}

To enable the model to gradually acquire and robustly exploit temporal information, we further design a coarse-to-fine training strategy, including temporal-aware finetuning and temporal-aware reinforcement learning.

\vspace{-4mm}
\paragraph{Stage 1: Temporal-aware Supervised Finetuning.}

In the first stage, we optimize the ChronusOmni via supervised fine-tuning (SFT). 
The model is trained on the video and audio dense captioning task, where for each video-audio pair input, the model learns to localize events and output timestamps, visual caption and audio caption for each event.

This training objective, together with our temporal-synchronized representation introduced in Section~\ref{temporal_modeling}, guides the model to understand the alignment relationships among video, audio, and time.

However, SFT alone presents inherent limitations for temporal grounding~\cite{timer1}. First, although time is a continuous variable, SFT treats it as a categorical label, ignoring the metric structure—i.e., the magnitude and distance between time points.
Second, as SFT exposes the model to gold prefixes throughout training, the model tends to rely on pattern memorization rather than learning to precisely localize boundaries.
Consequently, SFT struggles to establish a robust mapping between audiovisual features and accurate temporal intervals.
\vspace{-4mm}
\paragraph{Stage 2: Reinforcement Learning with GRPO.}
To overcome the limitations of SFT, we employ reinforcement learning in the second stage. 
Specifically, we adopt Group Relative Policy Optimization (GRPO)~\cite{deepseekmath,deepseekr1} to train the model on the audiovisual temporal grounding task directly.
This paradigm replaces the restrictive maximum likelihood objective with task-specific, outcome-driven reward functions, enabling the model to explore better strategies for aligning and integrating detailed audiovisual information with temporal boundaries.

For moment retrieval subtasks (V2T, A2T), the model outputs a temporal interval. We adopt the Intersection over Union (IoU) between the predicted interval and the ground-truth interval as the reward function:
\vspace{-1mm}
\begin{equation}
R_{\text{IoU}} = \frac{|I_{\text{pred}} \cap I_{\text{gt}}|}{|I_{\text{pred}} \cup I_{\text{gt}}|},
\vspace{-1mm}
\end{equation}
where $I_{\text{pred}}$ is the predicted time interval, and $I_{\text{gt}}$ is the ground-truth interval.
To encourage consistent formatting that aligns with our timestamp design (Section~\ref{temporal_modeling}), we also introduce a format reward enforcing outputs in the form ``second\{start\}-second\{end\}'':
\vspace{-1mm}
\begin{equation}
R_{\text{format}} = 
\begin{cases} 
0, & \text{if output format is wrong}, \\
1, & \text{if output format is correct}.
\end{cases}
\vspace{-1mm}
\end{equation}

For subtasks requiring the model to output a video or audio caption (T2A, V2A, T2V, A2V), we use Meteor~\cite{meteor} as the reward, which is a commonly used metric for evaluating video caption quality~\cite{usemeteor1,usemeteor2}:
\vspace{-1mm}
\begin{equation}
R_{\text{Meteor}} = \text{Meteor}(C_{\text{pred}},C_{\text{gt}}),
\vspace{-1mm}
\end{equation}
where $C_{\text{pred}}$ and $C_{\text{gt}}$ is the predicted  and ground-truth caption respectively.

This coarse-to-fine strategy effectively mitigates the discretization and exposure-bias issues of SFT by incorporating reward-based temporal supervision. As a result, the model achieves more accurate boundary localization and stronger audiovisual alignment, yielding substantially improved multimodal temporal grounding performance.

\section{ChronusAV: Multimodal Dataset for Audiovisual Temporal Grounding}

\begin{figure*}[t]
\centering
\includegraphics[width=1\linewidth]{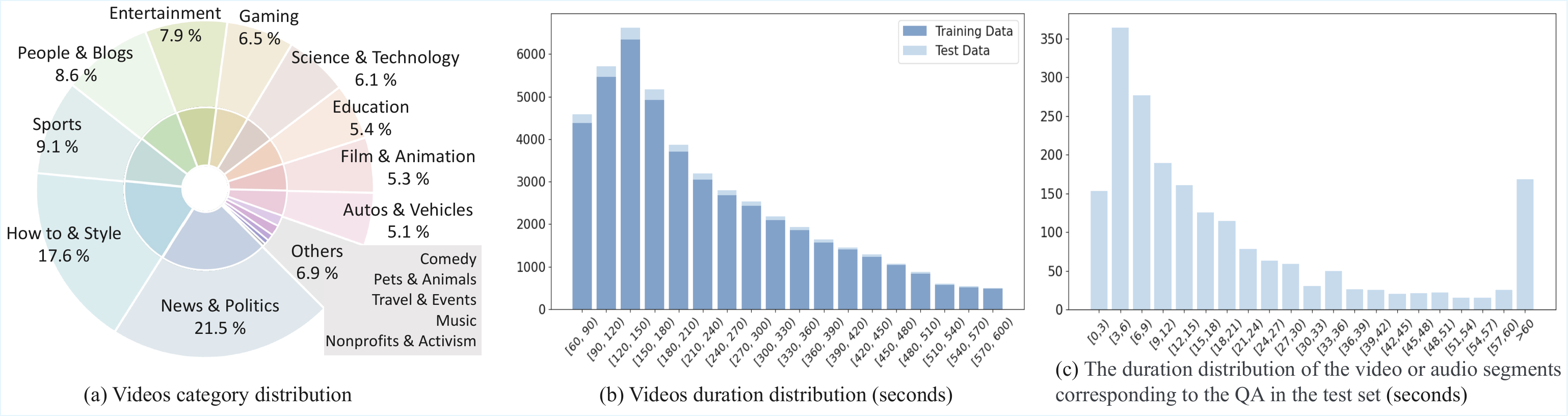} 
\vspace{-2 em}
\caption{Statistics of ChronusAV. 
}
\label{fig:dataset}
\vspace{-0.8 em}
\end{figure*}

\begin{table*}[t]
\centering
\caption{Comparison of ChronusAV to existing related datasets. ChronusAV covers large-scale open-domain long videos and, among the listed benchmarks, is the only dataset that simultaneously provides precise timestamps, multimodal annotations across vision, sound/music, and speech, as well as separate captions for audio and video. A\&V: audio and video. }
\vspace{-0.8 em}
\resizebox{\textwidth}{!}{ 
\begin{tabular}{l|ccccccccc}
\toprule

\textbf{Benchmark} & \textbf{Videos} & \textbf{Avg. video len} & \textbf{Annotations} & \textbf{Domain} & \textbf{Timestamps} & \textbf{Vision} & \textbf{Sound\&Music} & \textbf{Speech} & \textbf{A\&V seperate captions}\\

    \midrule  
Charades-STA\cite{gunnar2016charades} &10K&30s&16K&daily activity& \ding{51} & \ding{51} & \ding{55} & \ding{55} & -\\
ActivityNet Caps\cite{activitynet}  &20K &180s &72K & daily activity& \ding{51} & \ding{51} & \ding{55} & \ding{55} & -\\
VALOR~\cite{valor}   & 1.18M & 10s & 1.18M & open & \ding{55} & \ding{51} & \ding{51} & \ding{55} & \ding{55}\\
VAST~\cite{vast}    & 27M & 30s & 27M & open & \ding{55} & \ding{51} & \ding{51} & \ding{51} & \ding{55}\\
AVEL~\cite{AVEL} &4K & 10s &4K & open &\ding{51}& \ding{51}& \ding{51}& \ding{55}& \ding{55}\\
UnAV-100~\cite{Unav100}  & 30K & 42.1s &84K & open &\ding{51}& \ding{51}& \ding{51}& \ding{55}& \ding{55}\\
Shot2Story~\cite{shot2story} &43K&17.1s& 181K & open & \ding{51} & \ding{51}& \ding{55} & \ding{51} & \ding{51} \\
VUE-TR~\cite{vidi} & 428 & 907s  & 1598 &open & \ding{51} & \ding{51}& \ding{55} & \ding{51} & \ding{51} \\
LongVALE~\cite{longvale} &8.4K&235s&108K&open& \ding{51} & \ding{51}& \ding{51} & \ding{51} & \ding{55}\\
\midrule
ChronusAV  & 47K & 226s & 677K & open & \ding{51} & \ding{51}& \ding{51} & \ding{51} & \ding{51}\\
    \bottomrule
    \end{tabular}
}
\label{tab:benchmarks}
\vspace{-1.5 em}
\end{table*}

As discussed in Section \ref{sec:relatedworks}, to bridge the gap of missing audiovisual temporal grounding datasets, we construct ChronusAV. ChronusAV is a large-scale dataset tailored for the audiovisual temporal grounding task, containing separate, detailed captions for audio (including sound, speech, and music) and visual information with precise temporal boundaries.

\subsection{Construction of ChronusAV}
\label{sec:dataset_construction}
To construct a temporally-accurate, modality-complete, and cross-modal-aligned dataset, we design a systematic dataset construction pipeline:

\begin{itemize}
    \item \textbf{Video collection.} 
    To construct a real-world, open-domain dataset, we select English videos from the large-scale, high-resolution Panda-70M dataset~\cite{panda70m}. To cover videos of varying durations and reflect real-world temporal complexity, we select untrimmed long videos with audio tracks and restrict durations to 60–600 seconds, ensuring sufficient context for multimodal analysis.

\item \textbf{Video segmentation.} To obtain precise event timestamps, we segment long, untrimmed videos into meaningful and coherent event units. Following Panda-70M, we first split basic visual scenes and then merge semantically similar ones, ensuring that each final segment represents a complete and coherent event. The corresponding audio tracks are segmented using the same timestamps to maintain audio–visual alignment. Finally, we retain only those audio–video pairs with 5 to 30 segments, ensuring that each sample contains rich temporal dynamics without being excessively fragmented.

    \item \textbf{Modality-specific annotation.} 
    For each segment, we produce fine-grained, modality-aware captions by separately describing the visual and auditory streams. We use Gemini-2.5-Flash and Gemini-2.5-Pro~\cite{gemini25} to annotate video and audio, respectively. 

    \item \textbf{Human verification.}
    To ensure quality and modality separation of LLM-generated captions, we conduct a human study on 1,000 randomly sampled segments. Annotators rate accuracy and cross-modal leakage while viewing the corresponding video and audio. Results show that video captions are mostly accurate in 96.1\% of cases; audio captions in 93.5\%. And modality disentanglement is strong: 99.3\% of video captions and 97.5\% of audio captions show no or only minor cross-modal leakage. This high level of quality assurance validates the reliability of our automatically generated annotations for large-scale audiovisual temporal grounding training and evaluation.

    \item \textbf{Training set and benchmark construction.}
Ultimately, we annotate 677K segments from 47K untrimmed videos. Each annotation is a (timestamp, video caption, audio caption) triplet, supporting all subtasks of audiovisual temporal grounding.
We hold out 2,000 videos with corresponding audio to construct the test set and use the remaining 45K for training. For the test set, we randomly select one segment per video and generate six QA pairs covering the six subtasks, as shown in Figure \ref{fig:teaser}, yielding 12K QA pairs. All test QAs are human-verified and refined to ensure accurate, unique answers.

\end{itemize}
As a result, we get the ChronusAV, a temporally-accurate, modality-complete, and cross-modal-aligned dataset for audiovisual temporal grounding. 

\subsection{Analysis of ChronusAV}
As described in Section~\ref{sec:dataset_construction}, we construct the ChronusAV to enable the model training and a comprehensive evaluation of audiovisual temporal grounding. In this section, we provide an overview of the dataset and highlight the key characteristics that distinguish ChronusAV from existing temporal grounding datasets, which is also showed in Table~\ref{tab:benchmarks}.

\begin{itemize}
    \item \textbf{Large-scale and long-duration videos.} ChronusAV contains 47K multi-shot videos totaling 2,922 hours, with an average duration of 226 seconds.As shown in Figure~\ref{fig:dataset} (b), the videos range from 60 to 600 seconds, substantially longer than most prior benchmarks (e.g., ActivityNet Caps 180s, UnAV-100 42.1s). This long-form design provides rich temporal context and enables robust modeling of long-term event dependencies.
    \item \textbf{Diverse and open-domain coverage.} As shown in Figure~\ref{fig:dataset} (a), ChronusAV spans 15 real-world domains, covering a wide spectrum of scenarios. This domain diversity ensures that ChronusAV aligns with the distributions of natural audiovisual content and supports strong generalization across contexts.
    \item \textbf{Fine-grained temporal annotations.} ChronusAV provides precise timestamps, which are essential for temporal grounding tasks. In contrast, large-scale datasets like VALOR and VAST do not provide temporal boundaries, limiting their suitability for fine-grained temporal alignment or localization. In the test set, QA-related segments peak at 3–6 seconds, as shown in Figure~\ref{fig:dataset} (c), reflecting the benchmark’s 
    higher demands on a model’s ability to capture subtle temporal cues across modalities.
    \item \textbf{Modality completeness.} ChronusAV is one of the few datasets that simultaneously provide visual content and comprehensive audio cues (speech, music, and sound). Many existing datasets are audio-incomplete—e.g., Charades-STA and ActivityNet Caps provide no audio annotation; VALOR and AVEL lack speech; Shot2Story and VUE-TR lack music and sound.
    \item \textbf{Separate audio and video captions.} ChronusAV includes modality-separated audio and video captions, enabling detailed and disentangled multimodal understanding. Most existing datasets (including LongVALE) do not offer this feature. Although Shot2Story and VUE-TR provide separate captions, their annotations lack sound-event descriptions, limiting their ability to capture real-world auditory complexity.
\end{itemize}

These features makes ChronusAV can serve as a solid foundation for training and assessing our model. 
More statistics and cases can be found in Appendix.

\section{Experimental Settings}

\begin{table*}
  \centering
    \caption{Comparison with existing video-audio LLMs for multimodal temporal grounding task on ChronusAV benchmark. B: BLEU-4. R: ROUGE-L. M: METEOR. C: CIDEr. For Qwen3-Omni we use Qwen3-Omni-30B-A3B, while other models are all of size 7B.}
    \vspace{-2.5mm}
  \resizebox{1\linewidth}{!}{ 
  \begin{tabular}{l|cc|cccc|cc|cccc|cccc|cccc}
    \toprule
    \multirow{2}{*}{Model}  &
    \multicolumn{2}{c}{V2T} & \multicolumn{4}{c}{T2V} & \multicolumn{2}{c}{A2T} & \multicolumn{4}{c}{T2A} & \multicolumn{4}{c}{V2A}& \multicolumn{4}{c}{A2V}\\
    \cmidrule(lr){2-21} 
     &  R@0.5 & R@0.7 & B& R&M & C & R@0.5 & R@0.7 & B& R& M & C & B& R& M & C & B& R& M & C \\
    \midrule
    VideoLLaMA~\cite{videollama} &2.00&0.80&0.10&1.50&0.93&0.03 &1.70&0.80&0.05&1.13 & 0.63 &0.06 &0.08 &1.11&0.81 &0.06  &0.07 &1.22 &0.93 &0.00\\
    Ola~\cite{ola} &5.80&2.65&0.26&1.66&0.99&0.47 &5.50&2.55 &0.18&0.78&0.31&0.26 &0.15&1.15&0.53&0.45  &0.18 &1.60 &1.08& 0.40\\
    AVicuna~\cite{avicuna}  &10.75 &4.70  & 0.04 &1.03 &0.32 &0.12 &8.20 & 3.45  & 0.01 & 0.51 &0.13 &0.02 &0.02 &0.54 &0.23 &0.10 &0.11 &1.45 &0.65 & 0.38\\
    LongVALE-LLM~\cite{longvale} &9.50&3.65 &0.35&2.03&0.99&\underline{1.38} &4.25 & 1.25 &0.15 &1.31 &0.50 &0.19 & 0.10 &1.26 &0.64 &0.24 &0.21 &\underline{1.88} &1.02&\underline{0.87}\\
    Qwen2.5-Omni~\cite{qwen25omni} &7.05&3.00  &0.35 &1.89& 1.02 &0.79 &10.10 &3.70 &0.67 &1.01&0.64&0.26&\underline{0.54}&0.91&0.61&0.30 &0.19&1.61&0.97&0.58\\
    ARC-Hunyuan-Video~\cite{hunyuan}&36.10 &\underline{23.15} &0.24 &1.54&1.11&0.46 &36.85 &24.25 &0.18&1.17&0.65&0.37 & 0.12 & 0.94 & 0.70 &0.10 &0.11 &1.12 & 0.91 & 0.11\\  
    Qwen3-Omni~\cite{qwen3omni}  &\underline{37.85}&21.80 &\underline{0.37}&\underline{2.13}&\underline{1.62}&0.92 &\underline{46.70} &\underline{33.10} &\underline{0.94}&\underline{2.26}&\underline{1.20}&\underline{2.18} &0.35 &\underline{1.58} &\underline{1.25} &\underline{0.74} &\underline{0.22} &1.72&\underline{1.49}&0.39\\
    \midrule
    ChronusOmni  &\textbf{63.15} &\textbf{45.95}  & \textbf{1.16} &\textbf{3.37} & \textbf{2.12} & \textbf{5.07} &\textbf{90.50} & \textbf{79.85}  &\textbf{6.78}&\textbf{6.86}&\textbf{4.50}&\textbf{34.30}&\textbf{3.61}&\textbf{4.90}&\textbf{3.27}&\textbf{13.60} & \textbf{1.01} & \textbf{3.17} & \textbf{2.12 }& \textbf{3.03}\\
    \bottomrule
  \end{tabular}}
  \label{tab:compare_on_chronusbench}
  \vspace{-0.5 em}
\end{table*}

\begin{table*}
  \centering
    \caption{Comparison on LongVALE benchmark. ``*'' indicates that the model has been trained on the LongVALE training set.
    }
    \vspace{-2mm}
  \resizebox{0.85\linewidth}{!}{ 
  \begin{tabular}{l|cccc|ccc|cccc}
    \toprule
    \multirow{2}{*}{Model} & \multicolumn{4}{c}{Omni-TVG} & \multicolumn{3}{c}{Omni-DVC} & \multicolumn{4}{c}{Omni-SC}  \\
    \cmidrule(lr){2-12} 
     & R@0.3 & R@0.5 & R@0.7 & mIoU & SODA\_c & CIDEr & METEOR & BLUE-4 & ROUGE-L & CIDEr & METEOR \\
     \midrule
    VideoChat~\cite{videochat} & 2.2 & 0.9 & 0.4 & 3.0 & 0.7 & 0.2 & 0.9 & 0.5 & 9.6 & 0.0 & 8.2  \\
    VideoChatGPT~\cite{videochatgpt}  & 4.9 & 2.0 & 0.9& 5.0 & 0.7& 0.1& 0.9&0.4& 14.0& 0.9 & 5.9 \\
    VideoLLaMA~\cite{videollama}  & 2.5 & 1.1 & 0.3 & 1.9 & 0.6& 0.6& 0.9& 0.9& 11.5& 0.1& 8.9 \\
    PandaGPT~\cite{pandagpt}  &2.5& 1.0& 0.3 & 2.2 & 0.5 & 0.0 & 0.6 & 0.6& 14.9 & 0.3 & 8.9 \\
    NExT-GPT~\cite{nextgpt} & 4.3 & 1.9 & 0.7 & 4.0 &0.2 & 0.1& 0.3 & 0.4 & 10.2& 0.0 & 8.1 \\
    TimeChat~\cite{timechat}  & 5.8 & 2.6 & 1.1 & 5.2 & 1.6 & 0.1 & 1.4 &1.2 & 16.1 & 1.6 & 10.0 \\
    VTimeLLM~\cite{huang2024vtimellm}  & 7.5 & 3.4 & 1.3 & 6.4 & 2.4 & 0.2& 2.0 &1.0&14.5&1.6&5.5  \\
    TriSense~\cite{trisense}  & 14.8 & \underline{9.3} & \underline{4.7} & \underline{11.2}& -& -& -& 4.8&  21.9 & \underline{18.8} & 10.4\\
    LongVALE-LLM\raisebox{0ex}{\scalebox{1.2}{*}}~\cite{longvale}  & \underline{15.7}&8.6&3.9&11.0&\underline{2.8}&\textbf{7.9}&\underline{4.7}&\textbf{5.6}&\textbf{22.4}&\textbf{20.3}&\underline{10.9}\\
    \midrule
    ChronusOmni  & \textbf{49.7} & \textbf{32.5} & \textbf{17.6} & \textbf{34.5} & \textbf{3.7} & \underline{5.6} & \textbf{5.2} &\underline{5.5} &\underline{22.0} &\textbf{20.3} &\textbf{11.7}\\
    \bottomrule
  \end{tabular}}
  \label{tab:compare_on_longvale}
  \vspace{-3mm}
\end{table*}

\subsection{Implementation Details}
We adopt multimodal LLM Ola~\cite{ola} as our backbone, which is built upon Qwen-2.5-7B~\cite{qwen25}. 
For visual processing, we use OryxViT~\cite{oryx} as the vision encoder and uniformly sample 64 frames from each input video. On the audio side, we employ Whisper-Large-V3~\cite{whisperv3} as the speech encoder and BEATs~\cite{beats}as the sound and music encoder. To achieve a comprehensive representation of the audio content, the embedding features from  Whisper-Large-V3 and BEATs are concatenated across the channel dimension. The resulting audio features are then downsampled by a factor of 10 to reduce token length while preserving semantic information. Finally, both visual and audio features are projected through two-layer multi-layer perceptron adapters, which transform them into unified tokens for the LLM decoder.

\subsection{Training Settings}
We initialize \ModelName~with the pretrained Ola model and only finetune the LLM parameters, keeping all encoders and adapters frozen throughout training.

\begin{itemize}
\item\textbf{SFT stage.} In this stage, we train on a mixture of How2~\cite{how2}, AVSD~\cite{AVSD}, and ChronusAV datasets. We randomly sample 10K clips of How2-300h and train on the AVSR task. For AVSD, we split each dialogue into turn-level QA in every dialogue, randomly sample 30K QA pairs, and train on the AVQA task. For ChronusAV, we randomly sample 30K videos from
the training set and train on the dense video and audio caption task. We train on all 70k samples for 1 epoch.
\item\textbf{GRPO stage.} We construct the GRPO data using the ChronusAV training set. For each untrimmed video, we randomly select one segment and form a QA pair corresponding to one of the six subtasks using the segment's timestamp and captions.
In this stage, we ultimately use 4000 pairs of QA data to train for 1000 steps. 
\end{itemize}

\subsection{Evaluation Settings}
We evaluate ChronusOmni across a broad set of tasks and datasets to assess both its fine-grained temporal grounding ability and general audiovisual understanding.

\begin{itemize}
\item\textbf{ChronusAV.} We comprehensively assess ChronusOmni's temporal grounding performance using ChronusAV benchmark, and compare it with previous audiovisual LLMs.
For V2T and A2T subtasks, we follow the evaluation metrics for the moment retrieval task used in previous works~\cite{guo2025trace,timechat}, reporting Recall@1 at IoU thresholds of {0.5,0.7}. For another four subtasks, we use BLEU-4~\cite{bleu}, ROGUE-L~\cite{rouge}, METEOR~\cite{meteor}, and CIDEr~\cite{cider} for standard caption quality evaluation. 
\item\textbf{LongVALE.}
LongVALE~\cite{longvale} comprises three omni temporal tasks: Omni-TVG (predict the time segment corresponding to an omni event caption), Omni-DVC (predict timestamps and captions of all omni events), and Omni-SC (generate the omni caption for a specified time segment). We evaluate ChronusOmni on the LongVALE test set in a zero-shot setting.

\item\textbf{Charades-STA and ActivityNet.}
Charades-STA~\cite{gunnar2016charades} and ActivityNet~\cite{activitynet} are widely used video temporal grounding datasets. Because some of our training data (e.g., AVSD) contains Charades-STA videos, a fair zero-shot evaluation on Charades-STA is infeasible; thus, we fine-tune for 1 epoch on Charades-STA using GRPO and report results under fine-tuning. For ActivityNet, we evaluate zero-shot on the test set. To enable fair comparison with prior VLMs, we use video-only inputs (no audio) for both datasets during training and testing.
\item \textbf{General video and audio understanding benchmarks.} To evaluate whether our temporal modeling affects general audiovisual reasoning, we conduct zero-shot evaluations on four widely used benchmarks: Video-MME~\cite{videomme}, Librispeech~\cite{librispeech}, VisSpeech~\cite{visspeech}, and MUSIC-AVQA~\cite{musicavqa}. Video-MME is vision-only, Librispeech is audio-only, while VisSpeech and MUSIC-AVQA utilize both visual and audio modalities. 

\end{itemize}

\section{Experimental Results}

\subsection{Results on Audiovisual Temporal Grounding} 
\paragraph{Performance on ChronusAV.}
We compare ChronusOmni with other omni LLMs on ChronusAV benchmark.
As shown in Table~\ref{tab:compare_on_chronusbench}, ChronusOmni consistently outperforms existing audiovisual LLMs across all subtasks by more than 30\%. In V2T subtask, ChronusOmni reaches R@0.5/0.7 of 63.15/45.95, yielding 67\%/98\% gains over the next best model. In A2T subtask, the R@0.5/R@0.7 achieves 90.50/79.85, yielding 94\%/142\% improvements. For the remaining four subtasks, ChronusOmni delivers multiplicative gains across most captioning metrics, with particularly large magins on CIDEr, demonstrating stronger temporal audiovisual alignment.
ChronusOmni also significantly outperforms recent omni-modal models such as ARC-Hunyuan-Video~\cite{hunyuan} and Qwen3-Omni~\cite{qwen3omni}, especially on audio-related subtasks, highlighting the limitations of prior models in handling complex auditory temporal cues.
Notably, many baselines use more densely sampled video frames (e.g., 100 in LongVALE-LLM, 150 in ARC-Hunyuan-Video, 2 fps in Qwen3-Omni), yet ChronusOmni—using 64 frames—achieves superior temporal understanding, indicating both higher accuracy and greater efficiency.

\begin{table}
  \centering
    \caption{Comparison with VLMs capable of temporal grounding on Charades-STA and ActivityNet. FT: fine-tuned setting; ZS: zero-shot setting. ``*" indicates that this model is smaller in size than ChronusOmni, while the sizes of the other models are all 7B.}
    \vspace{-2mm}
  \resizebox{1\linewidth}{!}{ 
  \begin{tabular}{l|ccc|ccc}
    \toprule
        \multirow{2}{*}{Model} &
    \multicolumn{3}{c}{Charades-STA (FT)} & \multicolumn{3}{c}{ActivityNet (ZS)}\\
     \cmidrule(lr){2-7} 
   &  R@0.3 & R@0.5 & R@0.7 & R@0.3 & R@0.5 & R@0.7\\
    \midrule
    TimeChat~\cite{timechat}  &-&46.7&23.7&36.2&20.2&9.5\\
    Hawkeye~\cite{HawkEye}   & 72.5&58.3&28.8 &49.1&29.3&10.7\\
    VTimeLLM~\cite{huang2024vtimellm} &-&-&-&  44.0 &27.8& 14.3\\
    VTG-LLM~\cite{guo2025vtgllm}  &-&57.8&33.9&-&-&-\\
    TRACE~\cite{guo2025trace}  &-&61.7&41.4& { -} & { -} & { -}\\
    TimeSuite~\cite{Zeng2025videochatt}  &79.4&67.1&43.0& { -} & { -} & { -}\\
    iMOVE\raisebox{0ex}{\scalebox{1.2}{*}}~\cite{imove} & 79.8 & 68.5 & 45.3 & 42.4 & 23.1 & 12.1\\
    VideoChat-R1~\cite{li2025videochatr1}   &-&71.7&\underline{50.2}& - & 33.4 & 17.7\\
    Time-R1\cite{timer1}  &\underline{82.9}&\underline{72.2}&50.1& \textbf{58.6} & \textbf{39.0} & 21.4\\
    \ModelName~  &\textbf{85.1}&\textbf{75.0}&\textbf{54.2}& \underline{56.9} & \underline{38.2} & \textbf{22.1}\\
    \bottomrule
  \end{tabular}}
  \label{tab:charade_and_anet}
  \vspace{-6mm}
\end{table}

\begin{table*}
  \centering
  \vspace{-0.5 em}
    \caption{Ablation study on our temporal interleaved tokenization method and training strategy. TIT: temporal interleaved tokenization.}
    \vspace{-0.5 em}
  \resizebox{1\linewidth}{!}{ 
  \begin{tabular}{l|cc|cccc|cc|cccc|cccc|cccc}
    \toprule
    \multirow{2}{*}{Model}  &
    \multicolumn{2}{c}{V2T} & \multicolumn{4}{c}{T2V} & \multicolumn{2}{c}{A2T} & \multicolumn{4}{c}{T2A} & \multicolumn{4}{c}{V2A}& \multicolumn{4}{c}{A2V}\\
    \cmidrule(lr){2-21} 
     &  R@0.5 & R@0.7 & B& R&M & C & R@0.5 & R@0.7 & B& R& M & C & B& R& M & C & B& R& M & C \\
    \midrule
    w/o TIT  & 16.05 & 6.80 &0.64 &2.51 &1.67 &1.76& 26.05 &13.00 & 2.43 & 3.29 &2.99 &2.94 &2.56&3.74&2.75&4.17&0.67&2.58&1.79&1.51\\
    w/o SFT &59.40&37.20 &0.53&2.39&1.81&0.69 &84.45 &69.25 &2.45&3.29&1.89&6.91 &2.12&3.15&1.96&5.32 &0.49&2.34&1.84 &0.43\\
    w/o GRPO & 30.00 &15.45 &0.46&2.53&1.07&2.37 &34.55 &19.00 &0.04&1.48&0.35&1.85 &0.54&1.40&0.73&1.24 &0.12&1.81&0.66&0.76\\
    \midrule
    ChronusOmni  &\textbf{63.15} &\textbf{45.95} & \textbf{1.16} &\textbf{3.37 }& \textbf{2.12} &\textbf{ 5.07} &\textbf{90.50} & \textbf{79.85 } &\textbf{6.78}&\textbf{6.86}&\textbf{4.50}&\textbf{34.30}&\textbf{3.61}&\textbf{4.90}&\textbf{3.27}&\textbf{13.60 }& \textbf{1.01} &\textbf{ 3.17} & \textbf{2.12 }& \textbf{3.03}\\
    \bottomrule
  \end{tabular}}
  \label{tab:ablation}
  \vspace{-4mm}
\end{table*}

\paragraph{Performance on LongVALE.}
Table~\ref{tab:compare_on_longvale} shows ChronusOmni's performance on LongVALE benchmark. As shown in the table, ChronusOmni delivers state-of-the-art performance on most evaluation metrics, while achieving the second-best results on the rest. In the Omni-TVG task, it achieves an mIoU of 34.5, nearly three times higher than the next best models—TriSense (11.2) and LongVALE-LLM (11.0)—demonstrating markedly stronger temporal localization ability. For Omni-DVC, ChronusOmni obtains the highest SODA\_c (3.7) and METEOR (5.2), and ranks second in CIDEr (5.6), behind only LongVALE-LLM, which is specifically trained on the LongVALE training set.
In Omni-SC, ChronusOmni achieves the best METEOR (11.7) and ties for the best CIDEr (20.3), while its BLEU-4 (5.5) and ROUGE-L (22.0) remain within negligible margins (0.1 and 0.4) of the leading scores.

\subsection{Results on Visual-only Temporal Grounding} 
Besides audiovisual temporal grounding, we also evaluate ChronusOmni on visual-only temporal grounding. As shown in Table~\ref{tab:charade_and_anet}, ChronusOmni achieves state-of-the-art performance on Charades-STA (fine-tuned), obtaining 85.1/75.0/54.2 R@0.3/0.5/0.7. These results outperform the next best models by +2.2 (vs. Time-R1 at R@0.3), +2.8 (vs. Time-R1 at R@0.5), and +4.0 (vs. VideoChat-R1 at R@0.7). On ActivityNet (zero-shot), ChronusOmni ranks second at R@0.3 (56.9) and R@0.5 (38.2), only slightly behind Time-R1 by 1.7 and 0.8, respectively. It achieves the best R@0.7 (22.1), surpassing Time-R1 by +0.7, indicating stronger high-precision temporal localization under zero-shot transfer. Overall, ChronusOmni consistently outperforms or matches strong baselines on visual-only temporal grounding benchmarks, demonstrating robust and generalizable temporal understanding capabilities.

\subsection{Results on General Video and Audio Understanding}
\begin{figure}[t]
  \centering
  \setlength{\abovecaptionskip}{1.0mm}
   \includegraphics[width=1\linewidth]{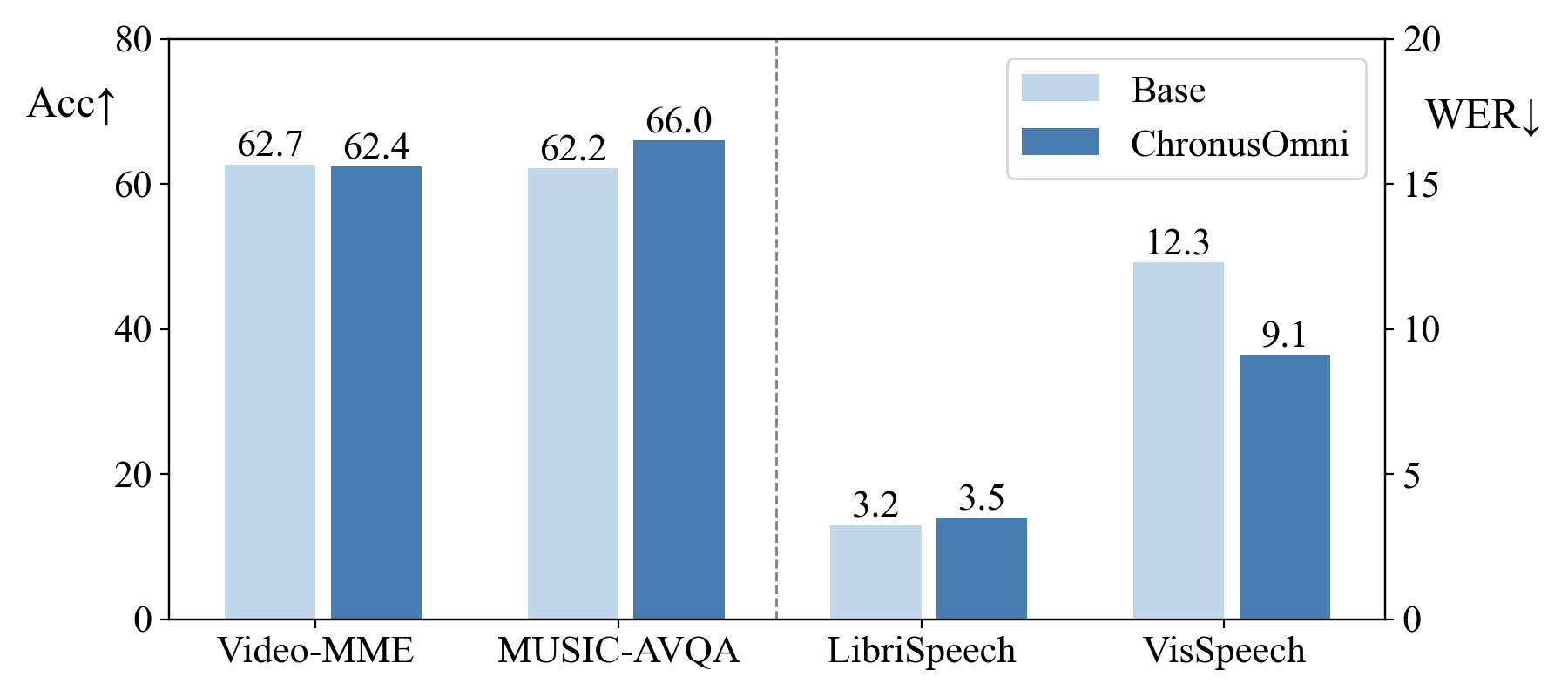}
   \caption{Evaluation on general video and audio understanding benchmarks. The evaluation metric for Video-MME and MUSIC-AVQA is Accuracy, with higher values being better. The evaluation metric for Librispeech and Visspeech is Word Error Rate (WER), with lower values being better. The ``Base" is Ola.}
   \label{fig:bar}
     \vspace{-6mm}
\end{figure}

We further evaluate whether audiovisual temporal-aware design affects the model’s general multimodal reasoning ability.
Figure~\ref{fig:bar} compares ChronusOmni with the base model across four standard video and audio understanding benchmarks. ChronusOmni matches the base model on general video QA (Video-MME) and shows only a minor degradation on pure speech recognition (LibriSpeech). In contrast, it yields substantial improvements on visual speech recognition (VisSpeech: 9.1 vs. 12.3 WER, a 26\% reduction) and audio-visual reasoning (MUSIC-AVQA: +3.8 over the base model).
It indicates that our temporal interleaved tokenization and temporal optimization do not impair general modality-specific abilities. Instead, they enhance the model’s ability to perform joint audio–visual reasoning, demonstrating stronger multimodal integration without sacrificing overall understanding quality.

\subsection{Ablation Study} 
Table~\ref{tab:ablation} shows the ablation results of our temporal interleaved tokenization method and training strategy.
Removing temporal interleaved tokenization leads to severe degradation in moment retrieval: V2T R@0.7 drops from 45.95 to 6.80 and A2T R@0.7 from 79.85 to 13.00, highlighting its essential role in absolute time perception. 
SFT is crucial for caption quality: T2V CIDEr rises from 0.69 (w/o SFT) to 5.07, and T2A from 6.91 to 34.30. 
GRPO provides the largest single gains in video-audio alignment: removing it reduces all V2A and A2V metrics by over 3 times. With all components enabled, ChronusOmni achieves the best results across all six directions and all metrics, confirming that all three components are complementary and jointly necessary for strong audiovisual multimodal temporal grounding.

\section{Conclusion} 
In this paper, we formally define audiovisual temporal grounding task and present \ModelName, a multimodal LLM tailored for this task. Our approach builds a temporally synchronized audio–visual representation and trains the model with temporal-aware supervised training and reinforcement learning to strengthen fine-grained temporal understanding. We also release ChronusAV, a comprehensive and standardized dataset for training and evaluation of audiovisual temporal grounding. Across ChronusAV and widely used public datasets, \ModelName~sets a new state of the art,  while maintaining robust video and audio understanding capabilities. These results demonstrate the effectiveness of our explicit cross-modal temporal alignment method and training strategy. Future work will focus on deploying our model in real-world interactive scenarios and scaling it to hour-long videos.

{
    \small
    \bibliographystyle{ieeenat_fullname}
    \bibliography{main}
}

\clearpage
\setcounter{page}{1}
\maketitlesupplementary

\section{Additional Details of ChronusAV Dataset}
\label{sec:ChronusAV}
\subsection{Additional Details of ChronusAV Construction}
Our dataset construction pipeline includes video collection, video segmentation, modality-specific annotation, human verification, and training set and benchmark construction, as introduced in Section 5.1. In this section, we provide further implementation details specifically focusing on the modality-specific annotation details and the rigorous human verification process employed to ensure data quality.

\begin{itemize}
\item \textbf{Details of modality-specific annotation.}
As introduced in Section 5.1, we use Gemini-2.5-Flash and Gemini-2.5-Pro to annotate video and audio segment caption, respectively. 

In video segment caption annotation, our prompt is: ``The user will input a video. Please provide a brief description of the main visual content of the video. Avoid specific timestamps and keep the content concise. Avoid using indicative phrases like `in the video' and directly output the visual information." For video segments that are 16 seconds or shorter, we sample frames at 1 fps; for video segments longer than 16 seconds, we uniformly sample 16 frames. We then input the sampled video frames along with the prompt into Gemini-2.5-Flash to obtain the output caption for each video segment. 

In audio segment caption annotation, our prompt is:
``The user will input a audio. Briefly describe the audio information, including the original text of the speech, audio events, etc. Avoid using indicative phrases like `in the audio' and directly output the audio information. Do not interpret the meaning of the speech expressed in the audio; just record what you hear concisely. For audio events, only output those that you are very certain about, and disregard any uncertain sounds. Record speech and audio events in order, but avoid specific timestamps. \#\#For example: A train whistle blows followed by a character speaking in a crisp voice, `Hello, my name is John. I would like to help you learn the numbers.' This is all accompanied by the sound of a train moving on its tracks." We input the audio segment with this audio caption prompt into Gemini-2.5-Pro to obtain the output caption for each audio segment.

\begin{figure}[t]
    \vspace{-1mm}
  \centering
  \setlength{\abovecaptionskip}{1.0mm}
   \includegraphics[width=1.0\linewidth]{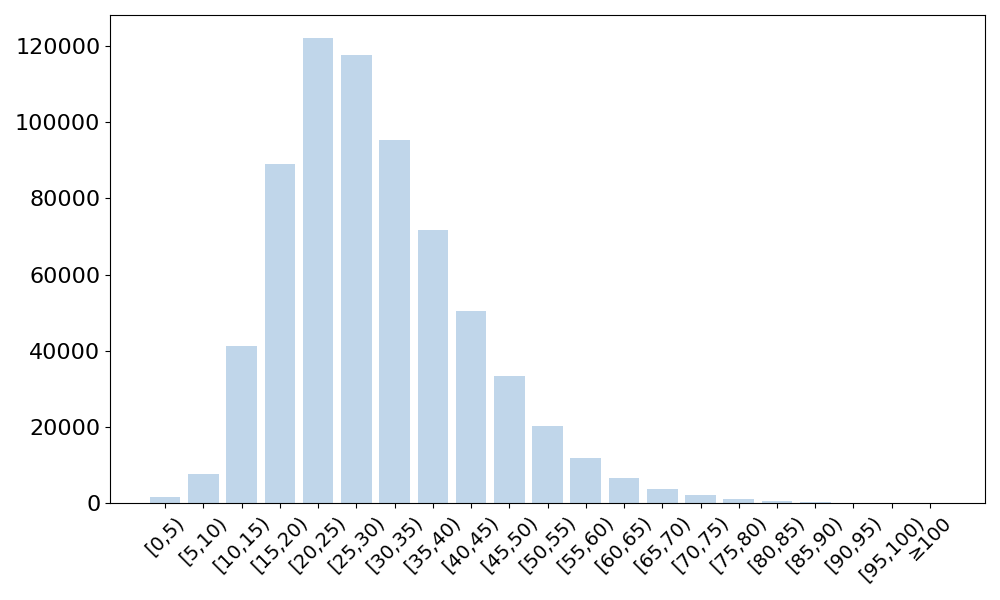}
    \vspace{-6mm}
   \caption{Distribution of visual segment caption length (words).}
   \label{fig:video_caption_length}
      \vspace{-6mm}
\end{figure}

\begin{figure}[t]
  \centering
  \setlength{\abovecaptionskip}{1.0mm}
   \includegraphics[width=1.0\linewidth]{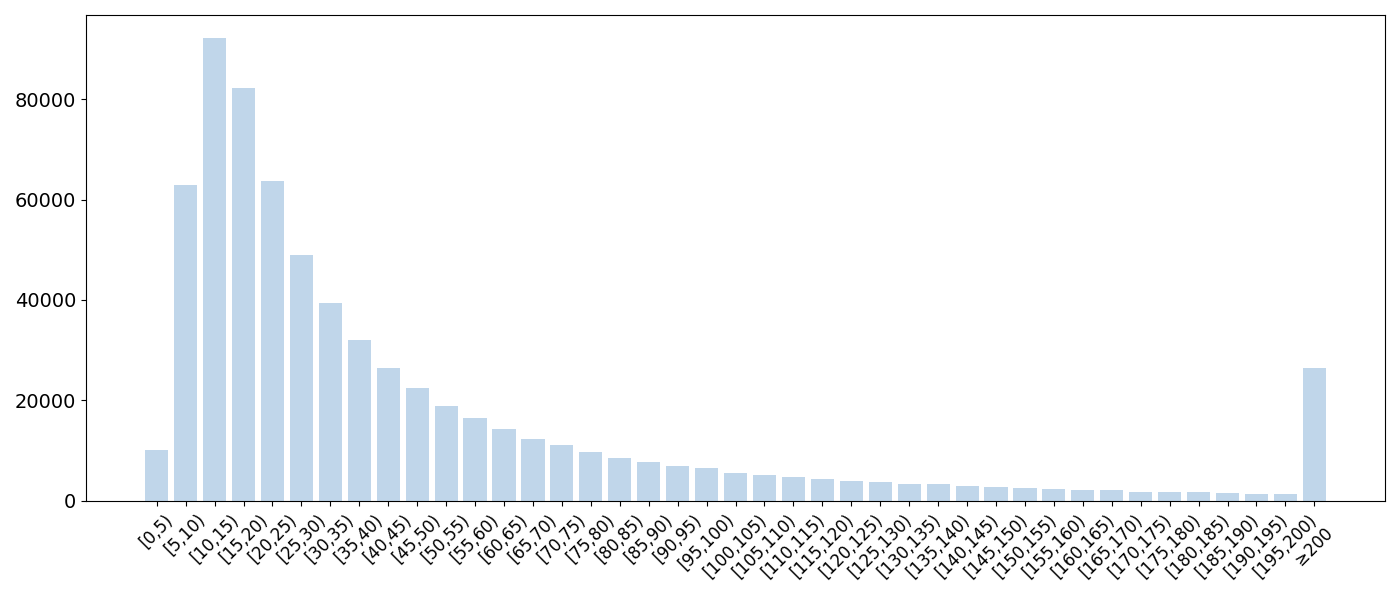}
   \caption{Distribution of audio segment caption length (words).}
   \label{fig:audio_caption_length}
   \vspace{-8mm}
\end{figure}

\item \textbf{Details of human verification.} 
To ensure quality of LLM-generated captions, we conduct a human study, as introduced in Section 5.1.
Three professional annotators are recruited to evaluate the annotations. They are presented with the sampled 1000 video and audio segments and asked to assess the generated captions based on two distinct criteria: \textit{Semantic Accuracy} and \textit{Modality Independence}.

For \textit{Semantic Accuracy}, we utilize a 3-point Likert scale: (1) \textit{Accurate}: The caption perfectly describes the visual/audio content;
(2) \textit{Acceptable}: The caption is generally correct but misses minor details or contains slight hallucinations;
(3) \textit{Inaccurate}: The caption is missing important details or contains significant hallucinations.

For \textit{Modality Independence} (Cross-modal Leakage), annotators checked for information leakage between modalities (e.g., visual captions describing auditory events like ``loud explosion''). This was rated as: (1) \textit{No Leakage}, (2) \textit{Minor Leakage}, or (3) \textit{Severe Leakage}.

To ensure consistency, we calculate the Fleiss' Kappa score among annotators, yielding $\kappa = 0.82$, indicating high inter-annotator agreement. Final labels are determined by majority voting.

\begin{figure*}[t]
\centering
\includegraphics[width=1\linewidth]{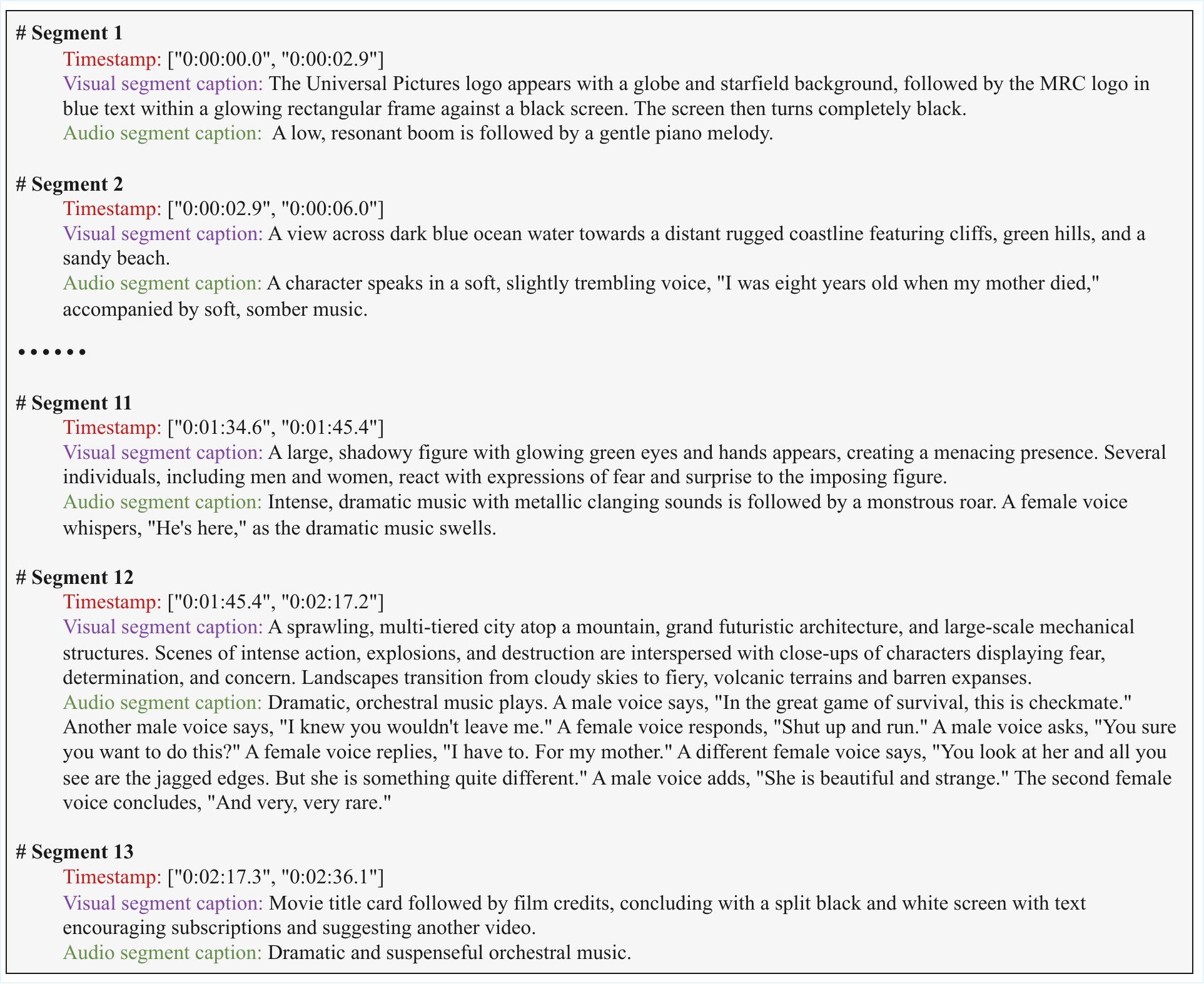} 
\caption{Timestamps, visual captions and audio captions of a 157-second video with corresponding audio in ChronusAV. 
}
\label{fig:datacase1}
\end{figure*}

\begin{figure*}[t]
\centering
\includegraphics[width=1\linewidth]{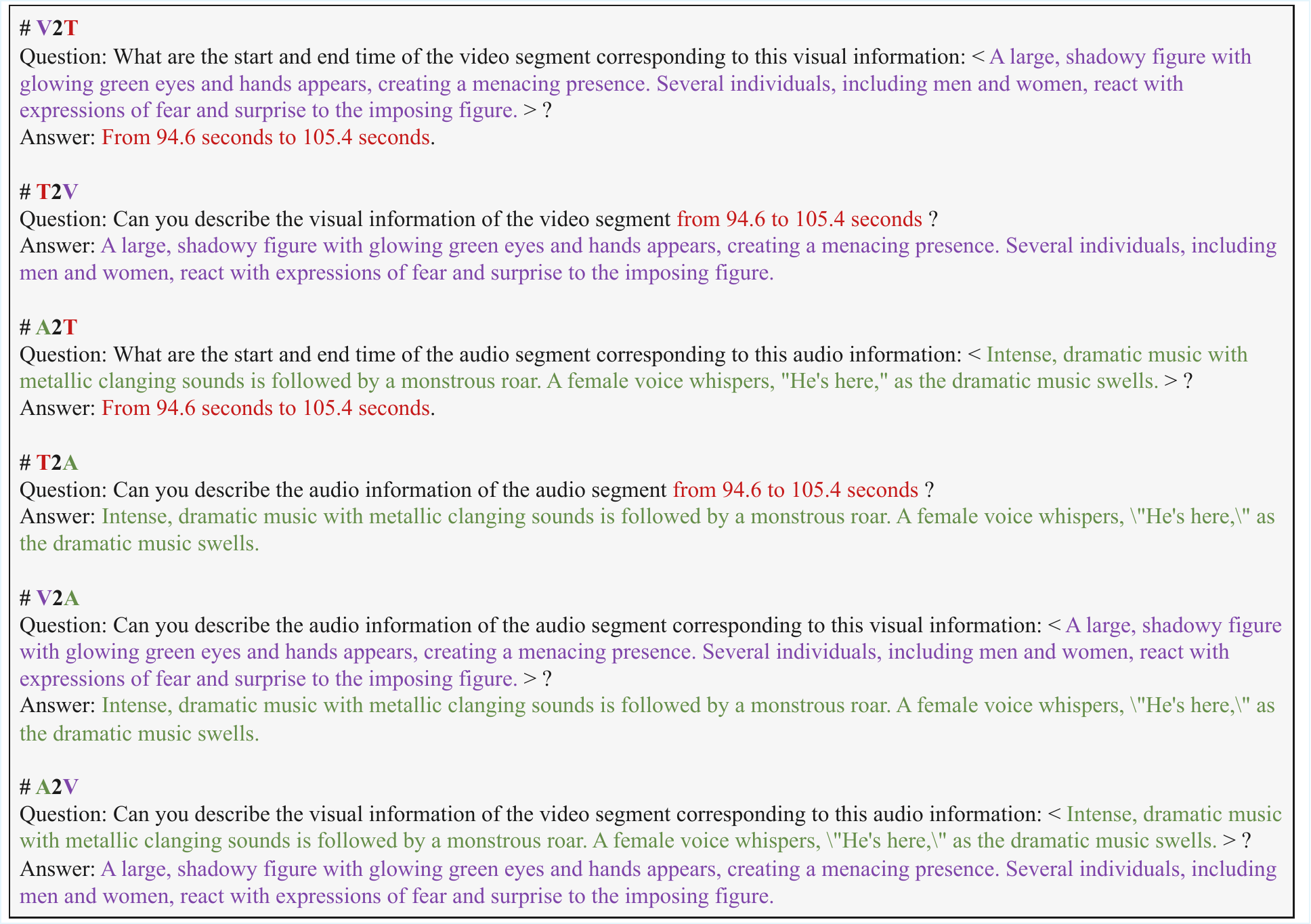} 
\caption{A case of question and answers for 6 subtasks in the audiovisual temporal grounding task.
}
\label{fig:datacase2}
     \vspace{-1mm}
\end{figure*}

Results show that video captions are mostly accurate (rated as Accurate or Acceptable) in 96.1\% of cases, and audio captions are mostly accurate in 93.5\%. Furthermore, 99.3\% of video captions and 97.5\% of audio captions show no or only minor cross-modal leakage. This high level of quality assurance validates the reliability of our automatically generated annotations for large-scale audiovisual temporal grounding training and evaluation.

\end{itemize}

\subsection{Additional Analyses of ChronusAV Dataset}
Figures~\ref{fig:video_caption_length} and Figures~\ref{fig:audio_caption_length} illustrate the distribution of caption lengths (in words) for visual and audio segments, respectively. As shown in Figure ~\ref{fig:video_caption_length}, the visual segment captions exhibit a unimodal, quasi-normal distribution. The vast majority of visual captions fall within the range of 15 to 40 words, with a distinct peak in the [20, 25) interval. The distribution tails off smoothly, with negligible instances of captions exceeding 80 words, suggesting a high degree of consistency in the length of visual descriptions. Figure ~\ref{fig:audio_caption_length} show that the audio segment caption length peaks in the [10, 20) interval, and exhibits a clear long-tail distribution. The long-tail distribution is due to the fact that some audio segments contains rich speech information and the detailed speech transcriptions significantly increase the length of the audio captions. The average length of visual segment captions is 29.6 words, while the average length of audio segment captions is 50.6 words, indicating that our captions are quite detailed.

A case showing the annotations of a 157-second video in ChronusAV, is presented in Figures~\ref{fig:datacase1}. For each untrimmed video with corresponding audio, we annotate the timestamps, visual segment captions and audio segment captions. Each (timestamp, visual segment caption, audio segment caption) tuple can be used to construct 6 types of QA in the audiovisual temporal grounding task. For example, we can use the timestamp, visual segment caption, audio segment caption of segment 11 in Figures~\ref{fig:datacase1} to construct 6 types of question-answer pairs, as shown in Figures~\ref{fig:datacase2}.

\section{Additional Details and Results of Experiments}
\label{sec:more_experiment}
\subsection{Training Hyperparameters}
In supervised fine-tuning stage, we train on 70K data samples for 1 epoch, as introduces in Section 6.2. We employ a batch size of 16. Optimization is guided by a cosine learning rate scheduler with a peak learning rate of $5 \times 10^{-6}$, a warmup ratio of 0.05, a minimum learning rate ratio of 0.01, and no weight decay. The maximum context length is configured to 32K tokens. We trained for about 17 hours using $8 \times $NVIDIA A800-80GB GPUs in this stage.

In reinforcement learning stage, we train on 4K data samples for 1 epoch, using a batch size of 4 for 1000 steps. 
During the rollout process within GRPO, the generation temperature is set to 1.0, and the maximum number of generated tokens is 1024. we sample 4 generations per prompt. The KL divergence penalty coefficient $\beta$ is set to 0.04 to ensure controlled deviation from the initial policy.
Learning rate is set to $1 \times 10^{-6}$. We trained for about 27 hours using $4 \times $NVIDIA A800-80GB GPUs in this stage.
\subsection{Additional Details of Evaluation Settings}

\begin{table*}
  \centering
    \caption{Ablation study on sampled frame number of ChronusOmni. B: BLEU-4. R: ROUGE-L. M: METEOR. C: CIDEr. }
  \resizebox{1\linewidth}{!}{ 
  \begin{tabular}{l|c|cc|cccc|cc|cccc|cccc|cccc}
    \toprule
    \multirow{2}{*}{Model}&\multirow{2}{*}{Frame Number}  &
    \multicolumn{2}{c}{V2T} & \multicolumn{4}{c}{T2V} & \multicolumn{2}{c}{A2T} & \multicolumn{4}{c}{T2A} & \multicolumn{4}{c}{V2A}& \multicolumn{4}{c}{A2V}\\
    \cmidrule(lr){3-22} 
     &  & R@0.5 & R@0.7 & B& R&M & C & R@0.5 & R@0.7 & B& R& M & C & B& R& M & C & B& R& M & C \\
    \midrule
    ChronusOmni &32 &51.40&32.95 &1.02&3.12&1.93&4.44 &83.85 &66.90 &6.32&6.36&4.22&27.02 &3.15&4.46&3.14&10.73&0.49&2.34&1.84 &0.43\\
    ChronusOmni &64 &\textbf{63.15} &\textbf{45.95} & 1.16 &3.37 & 2.12 &5.07 &\textbf{90.50} & 79.85&\textbf{6.78}&\textbf{6.86}&\textbf{4.50}&34.30&3.61&\textbf{4.90}&\textbf{3.27}&\textbf{13.60}& \textbf{1.01} &3.17 & \textbf{2.12} & 3.03\\
    ChronusOmni &128 & 60.20 &43.85 &\textbf{1.20}&\textbf{3.42}&\textbf{2.14}&\textbf{5.24} &90.30 &\textbf{82.90} &6.38&6.76&4.23&\textbf{34.62}&\textbf{3.62}&4.76&3.00&\textbf{13.60}
    &1.00&\textbf{3.19}&2.11&\textbf{3.43}\\
    \bottomrule
  \end{tabular}}
  \label{tab:ablation_frame_num}
\end{table*}

For all the models we evaluate, we use the officially recommended number of frames, for example, 8 frames for VideoLLaMA, 64 frames for Ola, 100 frames for Avicuna and LongVALE-LLM, 150 frames for ARC-Hunyuan-Video, 2 fps (for videos shorter than 384 seconds) or 768 frames (for videos longer than 384 seconds) for Qwen2.5-Omni and Qwen3-Omni. To ensure fairness and reproducibility, we use greedy sampling for all models on evaluation. Due to different models using various time formats during training (e.g. frame index for AVicuna and LongVALE-LLM, absolute time format ``HH:MM:SS" for ARC-Hunyuan-Video), we convert the time format in the ChronusAV benchmark to match the format used by the models during training in order to ensure the best inference results.

\subsection{Frame Sampling Ablations}
We perform an ablation study on the number of sampled frames during the inference stage, with the model trained on 64 frames. As shown in Table \ref{tab:ablation_frame_num}, the model achieves optimal performance across most subtasks when the inference frame number matches the training setting (64 frames). Reducing the frames to 32 results in a noticeable performance drop due to information loss. While increasing the sampling rate to 128 frames slightly improves some metrics by providing denser temporal guidance, it also degrades performance in some metrics. This decline is likely attributed to the gap between training and inference. Consequently, we maintain 64 frames as the default inference setting to ensure consistency and maximize performance.

\subsection{Inference Efficiency Evaluation}
Due to the introduction of additional time tokens and the token interleave process compared to the base model, We clearly compare the inference efficiency in Table \ref{tab:efficiency} to analyze the computational burden brought by our approach. We test the average inference CPU and GPU time of the base model and ChronusOmni over 2000 samples of the A2T subtask in ChronusAV benchmark on a single A800 GPU.
Compared with the base model, ChronusOmni introduces negligible computational overhead. Specifically, the average total inference time increases by only 0.21 seconds (from 3.52 s to 3.73 s), demonstrating that our method maintains the high efficiency of the backbone while incorporating additional capabilities.
\begin{table}
  \centering
    \caption{Comparison with base model on inference efficiency. ChronusOmni incurs only a slight increase in inference latency compared to the base model, striking a favorable balance between performance and efficiency.}
  \resizebox{1\linewidth}{!}{ 
  \begin{tabular}{l|ccc}
    \toprule
        \multirow{1}{*}{Model} &
   Avg. CPU Time &	Avg. GPU Time & Avg. Infer Time \\
    \midrule
 Base model (Ola) & 1.51 s & 2.01 s &3.52 s\\
 ChronusOmni & 1.55 s &2.18 s &3.73 s\\
    \bottomrule
  \end{tabular}}
  \label{tab:efficiency}
\end{table}

\subsection{Qualitative Results}
\begin{figure}[t]
  \centering
  \setlength{\abovecaptionskip}{1.0mm}
   \includegraphics[width=1.0\linewidth]{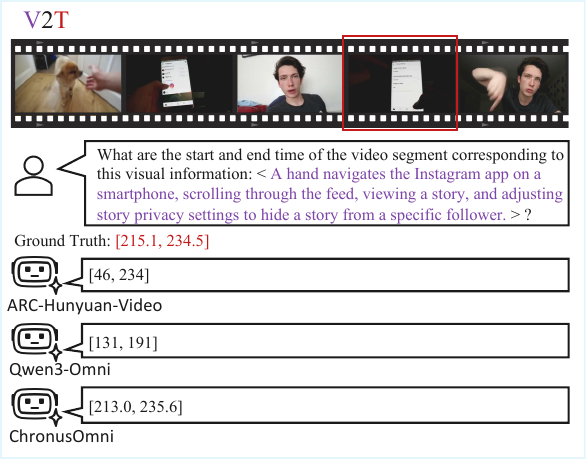}
   \caption{Qualitative results on V2T subtask. The sample is from ChronusAV benchmark.}
   \label{fig:v2t}
\end{figure}

\begin{figure}[t]
  \centering
  \setlength{\abovecaptionskip}{1.0mm}
   \includegraphics[width=1.0\linewidth]{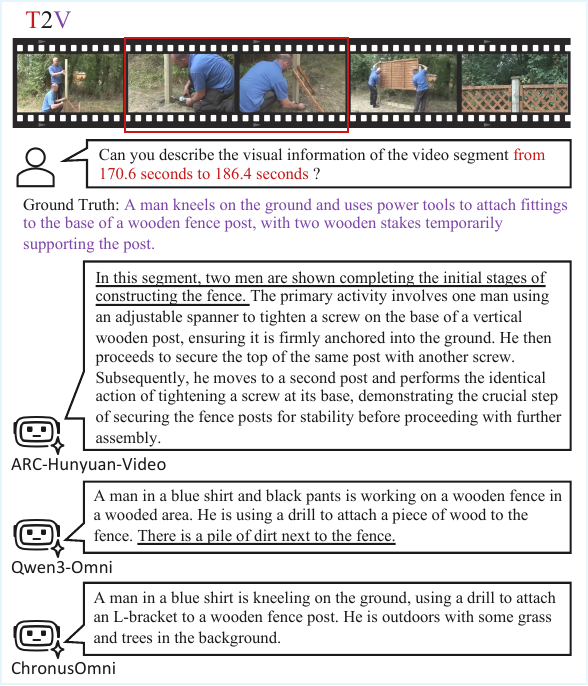}
   \caption{Qualitative results on T2V subtask. The sample is from ChronusAV benchmark. The underlined sentence indicates the incorrect part.}
   \label{fig:t2v}
\end{figure}

\begin{figure}[t]
  \centering
  \setlength{\abovecaptionskip}{1.0mm}
   \includegraphics[width=1.0\linewidth]{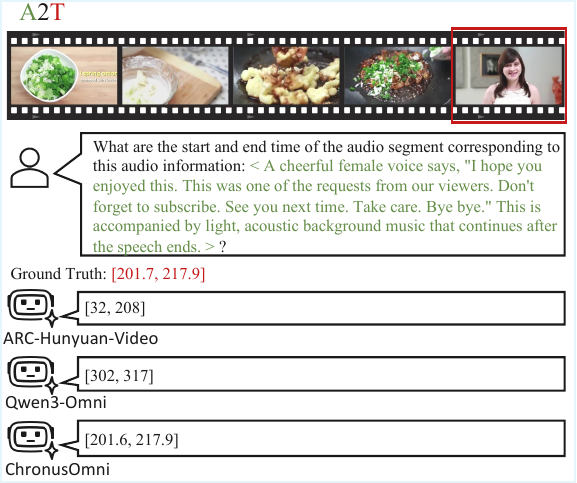}
   \caption{Qualitative results on A2T subtask. The sample is from ChronusAV benchmark.}
   \label{fig:a2t}
\end{figure}

\begin{figure}[t]
  \centering
  \setlength{\abovecaptionskip}{1.0mm}
   \includegraphics[width=1.0\linewidth]{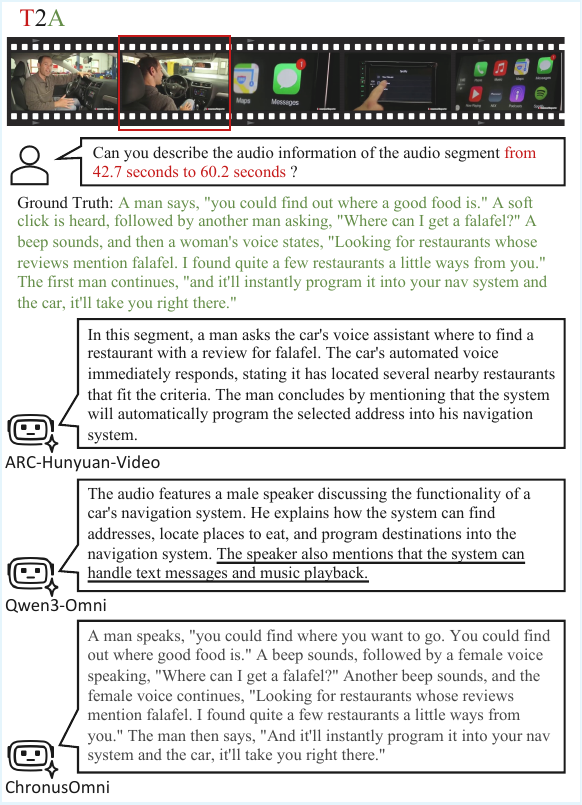}
   \caption{Qualitative results on T2A subtask. The sample is from ChronusAV benchmark. The underlined sentence indicates the incorrect part.}
   \label{fig:t2a}
\end{figure}

\begin{figure}[t]
  \centering
  \setlength{\abovecaptionskip}{1.0mm}
   \includegraphics[width=1.0\linewidth]{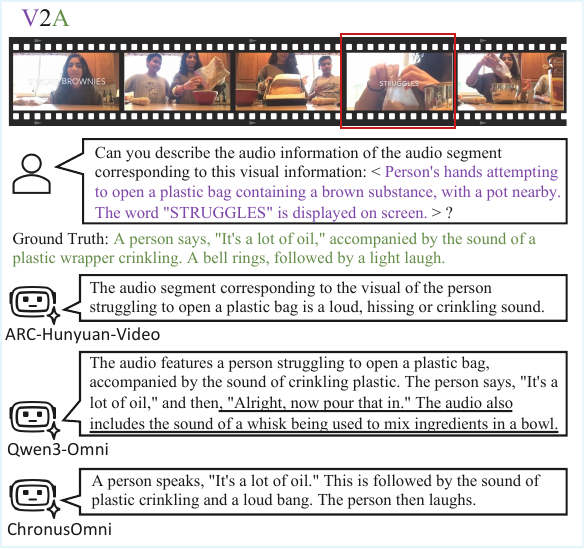}
   \caption{Qualitative results on V2A subtask. The sample is from ChronusAV benchmark. The underlined sentence indicates the incorrect part.}
   \label{fig:v2a}
\end{figure}

\begin{figure}[t]
  \centering
  \setlength{\abovecaptionskip}{1.0mm}
   \includegraphics[width=1.0\linewidth]{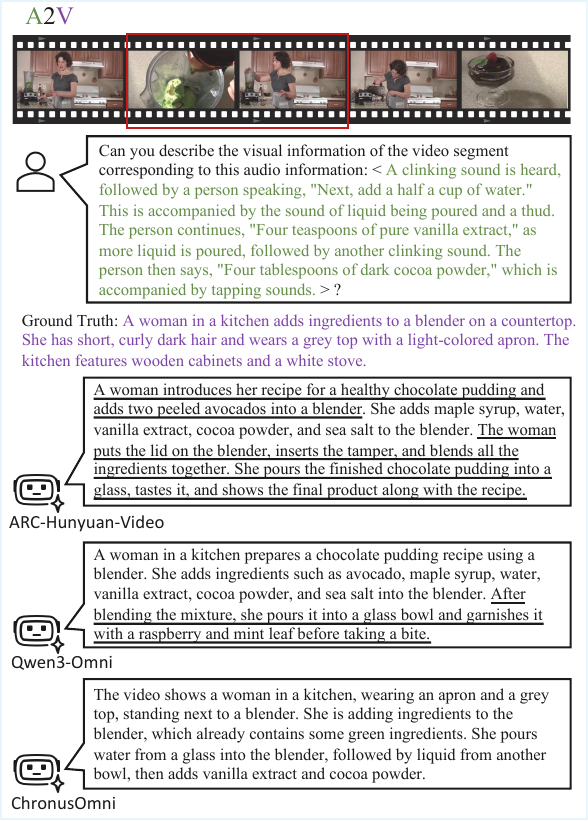}
   \caption{Qualitative results on A2V subtask. The sample is from ChronusAV benchmark. The underlined sentence indicates the incorrect part.}
   \label{fig:a2v}
\end{figure}

We provide qualitative results of ChronusOmni and other two audiovisual LLMs in various subtasks of ChronusAV benchmark in Figures~\ref{fig:v2t}-\ref{fig:a2v}. As shown in Figures~\ref{fig:v2t} and Figures~\ref{fig:a2t}, ChronusOmni can locate more precise time boundries. Figures~\ref{fig:t2v},~\ref{fig:t2a},~\ref{fig:v2a} and~\ref{fig:a2v} shows ARC-Hunyuan-Video and Qwen3-Omni often describe information outside of the time periods in the time-related caption subtasks (T2V, T2A, V2A, A2V), or they may omit information within the time periods. In comparison, ChronusOmni's output demonstrates stronger temporal awareness.

\end{document}